\documentclass[12pt,twoside]{article}
\usepackage{amsmath}
\usepackage{amssymb}
\usepackage{graphicx}
\makeindex

\def\,{\mskip 3mu} \def\>{\mskip 4mu plus 2mu minus 4mu} \def\;{\mskip 5mu plus 5mu} \def\!{\mskip-3mu}
\def\dispmuskip{\thinmuskip= 3mu plus 0mu minus 2mu \medmuskip=  4mu plus 2mu minus 2mu \thickmuskip=5mu plus 5mu minus 2mu}
\def\textmuskip{\thinmuskip= 0mu                    \medmuskip=  1mu plus 1mu minus 1mu \thickmuskip=2mu plus 3mu minus 1mu}
\textmuskip
\def\beq{\dispmuskip\begin{equation}}    \def\eeq{\end{equation}\textmuskip}
\def\beqn{\dispmuskip\begin{displaymath}}\def\eeqn{\end{displaymath}\textmuskip}
\def\bqa{\dispmuskip\begin{eqnarray}}    \def\eqa{\end{eqnarray}\textmuskip}
\def\bqan{\dispmuskip\begin{eqnarray*}}  \def\eqan{\end{eqnarray*}\textmuskip}

\newenvironment{keywords}{\centerline{\bf
Keywords}\vspace{0.5ex}\begin{quote}\small}{\par\end{quote}\vskip
1ex}
\def\subsection#1{{\vspace{1ex plus 1ex minus 0.5ex}\noindent\bf{#1.}}}
\def\nq{\hspace{-1em}}
\def\odt{{\textstyle{1\over 2}}}
\def\eps{\varepsilon}
\def\v{\boldsymbol}
\def\mean{\bar}
\def\eqsq{\simeq}
\def\p{{\scriptscriptstyle+}}
\def\pp{{\scriptscriptstyle++}}
\def\n{{n}}
\def\npp{\n}
\def\t{\pi}
\def\pin{{\scriptstyle\Pi}}
\def\Var{{\mbox{Var}}}
\def\Cov{{\mbox{Cov}}}
\def\qmbox#1{{\quad\mbox{#1}\quad}}
\def\SetR{{I\!\!R}}

\begin{document}

\title{\vskip -6mm\bf\Large\hrule height5pt \vskip 6mm
Distribution of Mutual Information from Complete and Incomplete Data\thanks{%
Preliminary results have been presented at the conferences NIPS-2001
\cite{Hutter:01xentropy} and UAI-2002 \cite{Hutter:02feature} and
KI-2003 \cite{Hutter:03mimiss}. This research was supported in
parts by the NSF grants 2000-61847 and 2100-067961.} \vskip
6mm \hrule height2pt \vskip 5mm}
\author{{\bf Marcus Hutter} and {\bf Marco Zaffalon}\\[3mm]
IDSIA, Galleria 2, CH-6928\ Manno (Lugano), Switzerland\\
\{marcus,zaffalon\}@idsia.ch \hspace{17ex} IDSIA-11-02}
\maketitle

\begin{abstract}
Mutual information is widely used, in a descriptive way, to
measure the stochastic dependence of categorical random variables.
In order to address questions such as the reliability of the
descriptive value, one must consider sample-to-population
inferential approaches. This paper deals with the posterior
distribution of mutual information, as obtained in a Bayesian
framework by a second-order Dirichlet prior distribution. The
exact analytical expression for the mean, and analytical
approximations for the variance, skewness and kurtosis are
derived. These approximations have a guaranteed accuracy level of
the order $O(n^{-3})$, where $n$ is the sample size. Leading order
approximations for the mean and the variance are derived in the
case of incomplete samples. The derived analytical expressions
allow the distribution of mutual information to be approximated
reliably and quickly. In fact, the derived expressions can be
computed with the same order of complexity needed for descriptive
mutual information. This makes the distribution of mutual
information become a concrete alternative to descriptive mutual
information in many applications which would benefit from moving
to the inductive side. Some of these prospective applications are
discussed, and one of them, namely \emph{feature selection}, is
shown to perform significantly better when inductive mutual
information is used.
\end{abstract}


\section{Introduction}\label{secInt}
Consider a data set of $n$ observations (or units) jointly
categorized according to the random variables $\imath$ and
$\jmath$, in $\{1,...,r\}$ and $\{1,...,s\}$, respectively. The
observed counts are $\v n:=(n_{11},\ldots ,n_{rs})$, with
$n:=\sum_{ij}n_{ij}$, and the observed relative frequencies are
$\hat{\v\t}:=(\hat\t_{11},\ldots,\hat\t_{rs})$, with
$\hat\t_{ij}:=n_{ij}/n$. The data $\v n$ are considered as a
sample from a larger population, characterized by the actual
chances $\v\t:=(\t_{11},\ldots,\t_{rs})$, which are the
population counterparts of $\hat{\v\t}$. Both $\hat{\v\t}$
and $\v\t$ belong to the $rs$-dimensional unit simplex.

We consider the statistical problem of analyzing the association
between $\imath$ and $\jmath$, given only the data $\v n$. This
problem is often addressed by measuring indices of independence,
such as the statistical coefficient $\phi^2$
\cite[pp.~556--561]{KenStu67}. In this paper we focus on the index
$I$ called \emph{mutual information} (also called \emph{cross
entropy} or \emph{information gain}) \cite{Kul68}. This index has
gained a growing popularity, especially in the artificial
intelligence community. It is used, for instance, in learning
\emph{Bayesian networks}
\cite{ChowLiu68,Pearl88,Buntine:96,Heckerman:99}, to connect
stochastically dependent nodes; it is used to infer classification
trees \cite{Quinlan93}. It is also used to select \emph{features}
for classification problems \cite{DuHaSt01}, i.e.\ to select a
subset of variables by which to predict the \emph{class} variable.
This is done in the context of a \emph{filter approach} that
discards irrelevant features on the basis of low values of mutual
information with the class
\cite{Lew92,BluLan97,CheHatHayKroMorPagSes02}.

Mutual information is widely used in descriptive rather than
inductive way. The qualifiers `descriptive' and `inductive' are
used for models bearing on $\hat{\v\t}$ and $\v\t$, respectively.
Accordingly, $\hat{\v\t}$ are called \emph{relative frequencies},
and $\v\t$ are called \emph{chances}. At descriptive level,
variables $\imath$ and $\jmath$ are found to be either independent
or dependent, according to the fact that the \emph{empirical}
mutual information $I(\hat{\v\t})$ is zero or is a positive
number. At inductive level, $\imath$ and $\jmath$ are assessed to
be either independent or dependent only with some probability,
because $I(\v\t)$ can only be known with some (second order)
probability.

The problem with the descriptive approach is that it neglects the
variability of the mutual information index with the sample, and
this is a potential source of fragility of the induced models. In
order to achieve robustness, one must move from the descriptive to
the inductive side. This involves regarding the mutual information
$I$ as a random variable, with a certain distribution. The
distribution allows one to make reliable, probabilistic statements
about $I$.

In order to derive the expression for the distribution of $I$, we
work in the framework of Bayesian statistics. In particular, we
use a second order prior distribution $p(\v\t)$ which takes into
account our uncertainty about the chances $\v\t$. From the prior
$p(\v\t)$ and the likelihood we obtain the posterior $p(\v\t|\v
n)$, of which the posterior distribution $p(I|\v n)$ of the mutual
information is a formal consequence.

Although the problem is formally solved, the task is not
accomplished yet. In fact, closed-form expressions for the
distribution of mutual information are unlikely to be available,
and we are left with the concrete problem of using the
distribution of mutual information in practice. We address this
problem by providing fast analytical approximations to the
distribution which have guaranteed levels of accuracy.

We start by computing the mean and variance of $p(I|\v\n)$. This
is motivated by the central limit theorem that ensures that
$p(I|\v\n)$ can be well approximated by a Gaussian distribution
for large $n$. Section \ref{secApprox} establishes a general
relationship, used throughout the paper, to relate the mean and
variance to the covariance structure of $p(\v\t|\v\n)$. By
focusing on the specific covariance structure obtained when the
prior over the chances is Dirichlet, we are then lead to
$O(n^{-2})$ approximations for the mean and the variance of
$p(I|\v n)$. Generalizing the former approach, in Section
\ref{secGeneral} we report $O(n^{-3})$ approximations for the
variance, skewness and kurtosis of $p(I|\v\n)$. We also provide an
exact expression for the mean in Section \ref{secExact}, and
improved tail approximations for extreme quantiles.

By an example, Section \ref{secNum} shows that the approximated
distributions, obtained by fitting some common distributions to
the expressions above, compare well to the ``exact'' one obtained by
Monte Carlo sampling also for small sample sizes. Section
\ref{secNum} also discusses the accuracy of the approximations and
their computational complexity, which is of the same order of
magnitude needed to compute the empirical mutual
information. This is an important result for the real application
of the distribution of mutual information.

In the same spirit of making the results useful for real
applications, and considered that missing data are a pervasive
problem of statistical practice, we generalize the framework to
the case of incomplete samples in Section \ref{secMD}. We derive
$O(n^{-1})$ expressions for the mean and the variance of
$p(I|\v\n)$, under the common assumption that data are
\emph{missing at random} \cite{LitRub87}. These expressions are in
closed form when observations from one variable, either $\imath$
or $\jmath$, are always present, and their complexity is the same
of the complete-data case. When observations from both $\imath$
and $\jmath$ can be missing, there are no closed-form expressions
in general but we show that the popular expectation-maximization
(EM) algorithm \cite{Chen:74} can be used to compute $O(n^{-1})$
expressions. This is possible as EM converges to the global
optimum for the problem under consideration, as we show in Section
\ref{secMD}.

We stress that the above results are a significant and novel step
to the direction of robustness. To our knowledge, there are only
two other works in literature that are close to the work presented
here. Kleiter has provided approximations to the mean and the
variance of mutual information by heuristic arguments
\cite{Kleiter:99}, but unfortunately, the approximations are shown
to be crude in general (see Section \ref{secMI}). Wolpert and Wolf
computed the exact mean of mutual information
\cite[Th.10]{WolWol95} and reported the exact variance as an
infinite sum; but the latter does not allow a straightforward
systematic approximation to be obtained.

In Section \ref{sec:discussion} we move from the theoretical to
the applied side, discussing the potential implications of the
distribution of mutual information for real applications. For
illustrative purposes, in the following Section \ref{secFS}, we
apply the distribution of mutual information to feature selection.
We define two new filters based on the distribution of mutual
information that generalize the traditional filter based on
empirical mutual information \cite{Lew92}. Several experiments on
real data sets show that one of the new filters is more effective
than the traditional one in the case of sequential learning tasks.
This is the case for complete data described in Section
\ref{DS}, as well as incomplete data in Section \ref{EAWIS}.
Concluding remarks are reported in Section \ref{C}.

\section{Expectation and Variance of Mutual Information}\label{secMI}

\subsection{Setup}
Consider discrete random variables $\imath\in\{1,...,r\}$ and
$\jmath\in \{1,...,s\}$ and an i.i.d.\ random process with outcome
$(i,j)\in\{1,...,r\}\times\{1,...,s\}$ having joint chance
$\t_{ij}$. The mutual information is defined by
\beq\label{mi}
  I({\v\t}) \;=\; \sum_{i=1}^r\sum_{j=1}^s
  \t_{ij}\ln{\t_{ij}\over\t_{i\p}\t_{\p j}} \;=\;
  \sum_{ij}\t_{ij}\ln\t_{ij} -
  \sum_{i}\t_{i\p}\ln\t_{i\p} -
  \sum_{j}\t_{\p j}\ln\t_{\p j},
\eeq
where $\ln$ denotes the natural logarithm and
$\t_{i\p}=\sum_j\t_{ij}$ and
$\t_{\p j}=\sum_i\t_{ij}$ are marginal chances.
Often the descriptive index $I(\hat{\v\t}) =
\sum_{ij}{\n_{ij}\over\npp} \ln{\n_{ij}\npp\over\n_{i\p}\n_{\p j}}$
is used in the place of the actual mutual information.
Unfortunately, the empirical index  $I(\hat{\v\t})$ carries no
information about its accuracy.
Especially $I(\hat{\v\t})\neq 0$
can have to origins; a true dependency of the random variables $\imath$
and $\jmath$ or just a fluctuation due to the finite sample size.
In the Bayesian approach to this problem one assumes a prior
(second order) probability density $p(\v\t)$ for the unknown
chances $\t_{ij}$ on the probability simplex. From this one can
determine the posterior distribution $p(\v\t|\v\n) \propto
p(\v\t)\prod_{ij}\t_{ij}^{\n_{ij}}$ (the $n_{ij}$ are
multinomially distributed). This allows to determine the
posterior probability density of the mutual information:%
\footnote{$I(\v\t)$ denotes the mutual information for the
specific chances $\v\t$, whereas $I$ in the context above is
just some non-negative real number. $I$ will also denote the
mutual information {\it random variable} in the
expectation $E[I]$ and variance $\Var[I]$. Expectations are
{\it always} w.r.t.\ to the posterior distribution
$p(\v\t|\v\n)$. }
\beq\label{midistr}
  p(I|\v\n) = \int
  \delta(I(\v\t)-I)p(\v\t|\v\n)d^{rs}\v\t.
\eeq
The $\delta()$ distribution restricts the integral to $\v\t$ for
which $I(\v\t)=I$. Since $0\leq I(\v\t)\leq I_{max}$ with sharp
upper bound $I_{max}:= \min\{\ln r,\ln s\}$, the domain of
$p(I|\v\n)$ is $[0,I_{max}]$, hence integrals over $I$ may be
restricted to such interval of the real line.

For large sample size, $p(\v\t|\v\n)$ gets strongly peaked around
$\v\t=\hat{\v\t}$ and $p(I|\v\n)$ gets strongly peaked around the
empirical index $I=I(\hat{\v\t})$. The mean $E[I] =
\int_0^\infty I\cdot p(I|\v\n)\,dI = \int
I(\v\t)p(\v\t|\v\n)d^{rs}\v\t$ and the variance
$\Var[I]=E[(I-E[I])^2]=E[I^2]-E[I]^2$ are of central interest.

\subsection{General approximation of expectation and variance of $I$}\label{secApprox}
In the following we (approximately) relate the mean and variance
of $I$ to the covariance structure of $p(\v\t|\v\n)$. Let
$\mean{\v\t}:=(\mean\t_{11},\ldots,\mean\t_{rs})$, with
$\mean\t_{ij}:=E[\t_{ij}]$. Since $p(\v\t|\v\n)$ is strongly
peaked around $\v\t=\hat{\v\t}\approx\mean{\v\t}$, for large
$\npp$ we may expand $I(\v\t)$ around $\mean{\v\t}$ in the
integrals for the mean and the variance. With
$\Delta_{ij}:=\t_{ij}-\mean\t_{ij}\in[-1,1]$ and using
$\sum_{ij}\t_{ij}= 1 =\sum_{ij}\mean\t_{ij}$ we get the following
expansion of expression (\ref{mi}):
\beq\label{miexp}
  I(\v\t) \;=\; I(\mean{\v\t}) +
  \sum_{ij}\ln\left({\mean\t_{ij}\over\mean\t_{i\p}\mean\t_{\p j}}\right)\Delta_{ij}
  + \sum_{ij}{\Delta_{ij}^2\over 2\mean\t_{ij}} -
  \sum_i{\Delta_{i\p}^2\over 2\mean\t_{i\p}} -
  \sum_j{\Delta_{\p j}^2\over 2\mean\t_{\p j}} +
  O(\Delta^3),
\eeq
where $O(\Delta^3)$ is bounded by the absolute value of (and
$\Delta^3$ is equal to) some homogenous cubic polynomial in the
$r\cdot s$ variables $\Delta_{ij}$. Taking the expectation, the
linear term $E[\Delta_{ij}]=0$ drops out. The quadratic terms
$E[\Delta_{ij}\Delta_{kl}] = \Cov_{(ij)(kl)}[\v\t]$ are the
covariance of $\v\t$ under $p(\v\t|\v\n)$ and they are
proportional to $\npp^{-1}$. Equation (\ref{mom3}) in Section
\ref{secGeneral} shows that $E[\Delta^3]=O(\npp^{-2})$, whence
\beq\label{exnlo}
  E[I] \;=\; I(\mean{\v\t}) + {1\over 2}
  \sum_{ijkl}\left({\delta_{ik}\delta_{jl}\over\mean\t_{ij}} -
  {\delta_{ik}\over\mean\t_{i\p}} -
  {\delta_{jl}\over\mean\t_{\p j}}\right)\Cov_{(ij)(kl)}[\v\t] +
  O(\npp^{-2}).
\eeq
The Kronecker delta $\delta_{ij}$ is $1$ for $i=j$ and $0$ otherwise.
The variance of $I$ in leading order in $\npp^{-1}$ is
\bqa\nonumber
  \Var[I] &=&
  E[(I-E[I])^2] \;\eqsq\;
  E\left[\left(\sum_{ij}\ln\left({\mean\t_{ij}\over
    \mean\t_{i\p}\mean\t_{\p j}}\right)\Delta_{ij}\right)^2\right]
  \;=\; \\\label{varlo}
  &=&
  \sum_{ijkl}\ln{\mean\t_{ij}\over\mean\t_{i\p}\mean\t_{\p j}}
  \ln{\mean\t_{kl}\over\mean\t_{k\p}\mean\t_{\p l}}
  \Cov_{(ij)(kl)}[\v\t],
\eqa
where $\eqsq$ denotes equality up to terms of order $\npp^{-2}$.
So the leading order term for the variance of mutual information
$I(\v\t)$, and the leading and second leading order term for the
mean can be expressed in terms of the covariance of $\v\t$ under
the posterior distribution $p(\v\t|\v\n)$.

\subsection{The (second order) Dirichlet distribution}\label{secDD}
Noninformative priors $p(\v\t)$ are commonly used if no explicit
prior information is available on $\v\t$. Most noninformative
priors lead to a Dirichlet posterior distribution $p(\v\t|\v\n)
\propto \prod_{ij}\t_{ij}^{\n_{ij}-1}$ with
interpretation\footnote{To avoid unnecessary
complications we are abusing the notation: $\n_{ij}$ is now the
sum of real \emph{and} virtual counts, while it formerly denoted
the real counts only. In case of Haldane's prior ($n''_{ij}=0$),
this change is ineffective.} $\n_{ij}=\n'_{ij}+\n''_{ij}$, where
the $\n'_{ij}$ are the number of outcomes $(i,j)$, and $\n''_{ij}$
comprises prior information
Explicit prior knowledge may also be specified by using virtual
units, i.e.\ by $n''_{ij}$, leading again to a Dirichlet
posterior.

The Dirichlet distribution is defined as follows:
\bqan
  p(\v\t|\v\n) &=&
  {1\over
  {\cal N}(\v\n)}\prod_{ij}\t_{ij}^{\n_{ij}-1}\delta(\t_\pp-1)
  \quad\mbox{with normalization}
  \\\label{norm}
  {\cal N}(\v\n) &=&
  \int\prod_{ij}\t_{ij}^{\n_{ij}-1}\delta(\t_\pp-1)
  d^{rs}\v\t \;=\;
  {\prod_{ij}\Gamma(\n_{ij})\over\Gamma(\npp)},
\eqan
where $\Gamma$ is the Gamma function.
Mean and covariance of $p(\v\t|\v\n)$ are
\beq\label{ecov}
  \mean\t_{ij} := E[\t_{ij}]=
  {\n_{ij}\over\npp} = \hat\t_{ij}, \quad
  \Cov_{(ij)(kl)}[\v\t] =
  {1\over\npp+1}(\hat\t_{ij}\delta_{ik}\delta_{jl}-
  \hat\t_{ij}\hat\t_{kl}).
\eeq

\subsection{Expectation and variance of $I$ under Dirichlet priors}\label{secEVIDD}
Inserting (\ref{ecov}) into (\ref{exnlo}) and (\ref{varlo}) we get,
after some algebra, the mean and variance of the mutual
information $I(\v\t)$ up to terms of order $\npp^{-2}$:
\bqa\label{exnlodi}\label{Jdef}
  E[I] & \;\simeq\; & J \;+\; {(r-1)(s-1)\over 2(\npp+1)},\qquad
  J \;:=\; \sum_{ij}{\n_{ij}\over\npp}\ln{\n_{ij}\npp\over
    \n_{i\p}\n_{\p j}} \;=\; I(\hat{\v\t}),
  \\ \label{varlodi}\label{Kdef}
  \nq\Var[I] & \;\simeq\; &
  {1\over\npp+1}(K-J^2), \qquad\qquad
  K \;:=\; \sum_{ij}{\n_{ij}\over\npp}\left(\ln{\n_{ij}\npp\over
    \n_{i\p}\n_{\p j}}\right)^2.
\eqa
$J$ and $K$ (and $L$, $M$, $P$, $Q$ defined later) depend on
$\hat\t_{ij}$ only, i.e.\ are $O(1)$ in $\v\n$. Strictly speaking
in (\ref{exnlodi}) we should make the expansion
${1\over\npp+1}={1\over\npp}+O(\npp^{-2})$, i.e.\ drop the
$+1$, but the exact expression (\ref{ecov}) for the covariance
suggests to keep it. We compared both versions with the ``exact''
values (from Monte-Carlo simulations) for various parameters
$\v\t$. In many cases the expansion in ${1\over\npp+1}$ was more
accurate, so we suggest to use this variant.

The first term for the mean is just the descriptive index
$I(\hat{\v\t})$. The second term is a correction, small when $\npp$
is much larger than $r \cdot s$.
Kleiter \cite{Kleiter:99} determined the correction by
Monte Carlo studies as $\min\{{r-1\over 2\npp},{s-1\over
2\npp}\}$. This is only correct if $s$ or $r$ is 2. The
expression $2E[I]/n$ he determined for the variance has a
completely different structure than ours.
Note that the mean is lower
bounded by ${const.\over\npp}+O(\npp^{-2})$, which is strictly
positive for large, but finite sample sizes, even if $\imath$ and
$\jmath$ are statistically independent and independence is
perfectly represented in the data ($I(\hat{\v\t})=0$). On the
other hand, in this case, the standard deviation
$\sigma=\sqrt{\Var[I]}\sim {1\over\npp}\sim E[I]$ correctly
indicates that the mean is still consistent with zero (where
$f\sim g$ means that $f$ and $g$ have the same accuracy, i.e.
$f=O(g)$ and $g=O(f)$).

Our approximations for the mean (\ref{exnlodi}) and variance
(\ref{varlodi}) are good if ${r \cdot s\over\npp}$ is small. For
dependent random variables, the central limit theorem ensures that
$p(I|\v\n)$ converges to a Gaussian distribution with mean $E[I]$
and variance $\Var[I]$. Since $I$ is non-negative it is more
appropriate to approximate $p(I|\v\t)$ as a Gamma ($=$ scaled
$\chi^2$) or a Beta distribution with mean $E[I]$ and variance
$\Var[I]$, which are of course also asymptotically correct.

\section{Higher Moments and Orders}\label{secGeneral}
A systematic expansion of all moments of $p(I|\v\n)$ to arbitrary
order in $\npp^{-1}$ is possible, but gets soon quite cumbersome.
For the mean we give an exact expression in Section
\ref{secFurther}, so we concentrate here on the variance, skewness
and kurtosis of $p(I|\v\n)$. The $3^{rd}$ and $4^{th}$ central
moments of $\v\t$ under the Dirichlet distribution are
\beq\label{mom3}
  E[\Delta_a\Delta_b\Delta_c] \;=\; {2\over(\npp+1)(\npp+2)}
  [2\hat\t_a\hat\t_b\hat\t_c
   - \hat\t_a\hat\t_b\delta_{bc}
   - \hat\t_b\hat\t_c\delta_{ca}
   - \hat\t_c\hat\t_a\delta_{ab}
   + \hat\t_a\delta_{ab}\delta_{bc}]
\eeq
\bqa
   E[\Delta_a\Delta_b\Delta_c\Delta_d] &=& {1\over\npp^2}
   [3\hat\t_a\hat\t_b\hat\t_c\hat\t_d
   - \hat\t_c\hat\t_d\hat\t_a\delta_{ab}
   - \hat\t_b\hat\t_d\hat\t_a\delta_{ac}
   - \hat\t_b\hat\t_c\hat\t_a\delta_{ad} \nq\\[-1ex]\nonumber
   && \qquad\qquad\qquad\; - \hat\t_a\hat\t_d\hat\t_b\delta_{bc}
   - \hat\t_a\hat\t_c\hat\t_b\delta_{bd}
   - \hat\t_a\hat\t_b\hat\t_c\delta_{cd} \nq\\\nonumber
   && \qquad\qquad\qquad\;
   + \hat\t_a\hat\t_c\delta_{ab}\delta_{cd}
   + \hat\t_a\hat\t_b\delta_{ac}\delta_{bd}
   + \hat\t_a\hat\t_b\delta_{ad}\delta_{bc}]
   +O(\npp^{-3})\nq
\eqa
with $a = ij$, $b = kl,...\in\{1,...,r\}\times\{1,...,s\}$ being
double indices, $\delta_{ab} = \delta_{ik}\delta_{jl},...$
$\hat\t_{ij}={\n_{ij}\over\npp}$. Expanding $\Delta^k =
(\t-\hat\t)^k$ in $E[\Delta_a\Delta_b...]$ leads to expressions
containing $E[\t_a\t_b...]$, which can be computed by a case
analysis of all combinations of equal/unequal indices $a,b,c,...$
using (\ref{norm}). Many terms cancel out leading to the above
expressions. They allow us to compute the order $\npp^{-2}$ term
of the variance of $I(\v\t)$. Again, inspection of (\ref{mom3})
suggests to expand in $[(\npp+1)(\npp+2)]^{-1}$, rather than in
$\npp^{-2}$. The leading and second leading order terms of the
variance are given below,
\bqa\label{var2ndo}
  \Var[I]
  &=& {K-J^2\over\npp+1} +
  {M+(r - 1)(s - 1)(\odt - J)-Q
  \over(\npp+1)(\npp+2)} + O(\npp^{-3})
  \\\label{Mdef}
  M &:=& \sum_{ij}
  \left({1\over\n_{ij}}-{1\over\n_{i\p}}-{1\over\n_{\p
  j}}+{1\over\npp}\right)
  \n_{ij}\ln{\n_{ij}\npp\over\n_{i\p}\n_{\p j}},
  \\\label{Qdef}
  Q &:=& 1-\sum_{ij}{\n_{ij}^2\over\n_{i\p}\n_{\p j}}.
\eqa
$J$ and $K$ are defined in (\ref{Jdef}) and (\ref{Kdef}). Note
that the first term ${K-J^2\over\n+1}$ also contains second order
terms when expanded in $\npp^{-1}$. The leading order terms for
the $3^{rd}$ and $4^{th}$ central moments of $p(I|\v\n)$ are
\beqn
  E[(I-E[I])^3] =
  {2\over\npp^2}[2J^3 - 3KJ + L] +
  {3\over\npp^2}[K + J^2 - P] +
  O(\npp^{-3}),\qquad\qquad
\eeqn
\beqn
  L := \sum_{ij}{\n_{ij}\over\npp}\left(\ln{\n_{ij}\npp\over
    \n_{i\p}\n_{\p j}}\right)^3\!\!\!,\quad
  P := \sum_i{\n(J_{i\p})^2\over\n_{i\p}} + \sum_j{\n(J_{\p j})^2\over\n_{\p j}},\quad
  J_{ij} := {\n_{ij}\over\npp}\ln{\n_{ij}\npp\over\n_{i\p}\n_{\p j}},
\eeqn
\beqn
  E[(I-E[I])^4]  =
  {3\over\npp^2}[K-J^2]^2 + O(\npp^{-3}),
  \qquad\qquad\qquad\qquad\qquad\qquad\qquad\quad
\eeqn
from which the skewness and kurtosis can be obtained by dividing
by $\Var[I]^{3/2}$ and $\Var[I]^2$, respectively. One can see that
the skewness is of order $\npp^{-1/2}$ and the kurtosis is
$3+O(\npp^{-1})$. Significant deviation of the skewness from $0$
or the kurtosis from $3$ would indicate a non-Gaussian $I$. These
expressions can be used to get an improved approximation for
$p(I|\v\n)$ by making, for instance, an ansatz
\beqn
  p(I|\v\n)\propto (1+\tilde b I+\tilde c I^2) \cdot p_0(I|\tilde\mu,\tilde\sigma^2)
\eeqn
and fitting the parameters $\tilde b$, $\tilde c$, $\tilde\mu$,
and $\tilde\sigma^2$ to the mean, variance, skewness, and kurtosis
expressions above. $p_0$ is any distribution with Gaussian limit.
>From this, quantiles $p(I > I_*|\v\n):=\int_{I_*}^\infty
p(I|\v\n)\, dI$, needed later (and in \cite{Kleiter:99}),
can be computed. A systematic expansion of arbitrarily high
moments to arbitrarily high order in $\npp^{-1}$ leads, in
principle, to arbitrarily accurate estimates (assuming convergence
of the expansion).

\section{Further Expressions}\label{secFurther}

\subsection{Exact value for $E[I]$}\label{secExact}
It is possible to get an exact expression for the mean mutual
information $E[I]$ under the Dirichlet distribution.
By noting that $x\ln x = {d\over d\beta}x^\beta|_{\beta=1}$,
($x = \{\t_{ij},\t_{i\p},\t_{\p j}\}$), one
can replace the logarithms in the last expression of
(\ref{mi}) by powers. From (\ref{norm}) we see that
$E[(\t_{ij})^\beta]={\Gamma(\n_{ij}+\beta)\Gamma(\npp)\over
\Gamma(\n_{ij})\Gamma(\npp+\beta)}$. Taking the
derivative and setting $\beta=1$ we get
\beqn
  E[\t_{ij}\ln\t_{ij}] = {d\over d\beta}E[(\t_{ij})^\beta]_{\beta=1}
  = {\n_{ij}\over\npp}[\psi(\n_{ij}+1)-\psi(\npp+1)].
\eeqn
The $\psi$ function has the following properties (see
\cite{Abramowitz:74} for details):
\beqn
  \psi(z)={d\ln\Gamma(z)\over dz}={\Gamma'(z)\over\Gamma(z)},\quad
  \psi(z+1)=\ln z + {1\over 2z} - {1\over 12z^2} + O({1\over z^4}),
\eeqn
\beq\label{psi2}
  \psi(n)=-\gamma+\sum_{k=1}^{n-1}{1\over k},\quad
  \psi(n+\odt)=-\gamma+2\ln 2+2\sum_{k=1}^n{1\over 2k-1}.
\eeq
The value of the Euler constant $\gamma$ is irrelevant here,
since it cancels out. Since the marginal distributions of
$\t_{i\p}$ and $\t_{\p j}$ are also Dirichlet (with parameters
$\n_{i\p}$ and $\n_{\p j}$), we get similarly
\bqan
  E[\t_{i\p}\ln\t_{i\p}] &=&
  {1\over\npp}\sum_i\n_{i\p}[\psi(\n_{i\p}+1)-\psi(\npp+1)],
  \\
  E[\t_{\p j}\ln\t_{\p j}] &=&
  {1\over\npp}\sum_j\n_{\p j}[\psi(\n_{\p j}+1)-\psi(\npp+1)].
\eqan
Inserting this into (\ref{mi}) and rearranging terms we get the
exact expression
\beq\label{miexex}
  E[I] = {1\over\npp}\sum_{ij}\n_{ij}
  [\psi(\n_{ij}+1)-\psi(\n_{i\p}+1)-\psi(\n_{\p
  j}+1)+\psi(\npp+1)].
\eeq
(This expression has independently been derived in \cite{WolWol95}
in a different way.) For large sample sizes, $\psi(z+1)\approx\ln
z$ and (\ref{miexex}) approaches the descriptive index
$I(\hat{\v\t})$ as it should. Inserting the expansion
$\psi(z+1)=\ln z+{1\over 2z}+...$ into (\ref{miexex}) we also get
the correction term ${(r-1)(s-1)\over 2\npp}$ of (\ref{exnlodi}).

The presented method (with some refinements) may also be used to
determine an exact expression for the variance of $I(\v\t)$. All but
one term can be expressed in terms of Gamma functions. The final
result after differentiating w.r.t.\ $\beta_1$ and $\beta_2$ can
be represented in terms of $\psi$ and its derivative $\psi'$. The
mixed term $E[(\t_{i\p})^{\beta_1}(\t_{\p j})^{\beta_2}]$ is more
complicated and involves confluent hypergeometric functions, which
limits its practical use \cite{WolWol95}.

\subsection{Large and small $I$ asymptotics}\label{secLASIA}
For extreme quantiles $I_* \approx 0$ or $I_* \approx I_{max}$,
the accuracy of the derived approximations in the last sections
can be poor and it is better to use tail approximations. In the
following we briefly sketch how the scaling behavior of
$p(I|\v\n)$ can be determined.

We observe that $I(\v\t)$ is small iff
$\t_{\imath\jmath}$ describes near independent random variables
$\imath$ and $\jmath$. This suggests the reparametrization
$\t_{ij} = \tilde\t_{i\p}\tilde\t_{\p j} + \Delta_{ij}$ in the
integral (\ref{midistr}). In order to make this representation
unique and consistent with $\t_\pp=1$, we have to restrict the $r
+ s + rs$ degrees of freedom ($\tilde\t_{i\p}, \tilde\t_{\p j},
\Delta_{ij}$) to $rs-1$ degrees of freedom by imposing $r+s+1$
constraints, for instance $\sum_i\tilde\t_{i\p} =
\sum_j\tilde\t_{\p j} = 1$ and $\Delta_{i\p} = \Delta_{\p j} = 0$
($\Delta_\pp = 0$ occurs twice). Only small $\v\Delta$ can lead to
small $I(\v\t)$. Hence, for small $I$ we may expand $I(\v\t)$ in
$\v\Delta$ in expression (\ref{midistr}). Inserting $\t_{ij} =
\tilde\t_{i\p}\tilde\t_{\p j} + \Delta_{ij}$ into (\ref{miexp}), we
get $I(\tilde\t_{i\p}\tilde\t_{\p j} + \Delta_{ij}) = \v\Delta^T
\v H(\tilde{\v\t})\v\Delta+O(\Delta^3)$ with %
$
  H_{(ij)(kl)} = \odt[\delta_{ik}\delta_{jl}/\tilde\t_{ij} -
               \delta_{ik}/\tilde\t_{i\p} -
               \delta_{jl}/\tilde\t_{\p j}]
$ %
(cf.\ (\ref{exnlo})) and $\v H$ and $\v\Delta$ interpreted as
$rs$-dimensional matrix and vector. $\v\Delta^T \v
H(\tilde{\v\t})\v\Delta = I$ describes an $rs$-dimensional
ellipsoid of linear extension $\propto\sqrt I$. Due to the $r+s-1$
constraints on $\v\Delta$, the $\v\Delta$-integration is actually
only over, say, $\v\Delta_\bot$ and $\v\Delta^T_\bot \v
H_\bot(\tilde{\v\t})\v\Delta_\bot = I$ describes the surface of a
$\bar d := (r-1)(s-1)$-dimensional ellipsoid only. Approximating
$p(\v\t|\v\n)$ by $p(\tilde{\v\t}|\v\n)$ in (\ref{midistr}), where
$\tilde\t_{ij}=\tilde\t_{i\p}\tilde\t_{\p j}$ we get
\beqn
  p(I|\v\n) \;=\; B(\v\n)\cdot I^{{\bar d\over 2}-1}
  + o(I^{{\bar d\over 2}-1})
  \qmbox{with}
  B(\v\n)=\int S_\bot(\tilde{\v\t}) p(\tilde{\v\t}|\v\n)
  d^{r+s-2}\tilde{\v\t}
\eeqn
where $S_\bot = {\pin^{\bar d/2}\over\Gamma(\bar d/2)\sqrt{\det{\v
H_\bot}}}$ is the ellipsoid's surface ($\pin=3.14...$). Note
that $d\tilde{\v\t}$ still contains a Jakobian from the non-linear
coordinate transformation. So the small $I$ asymptotics is
$p(I|\v\n)\propto I^{{\bar d\over 2}-1}$ (for any prior), but a
closed form expression for the coefficient $B(\v\n)$ has yet to be
derived.

Similarly we may derive the scaling behavior of $p(I|\v\n)$ for $I
\approx I_{max} := \min\{\ln r,\ln s\}$. $I(\v\t)$ can be written
as $H(\imath) - H(\imath|\jmath)$, where $H$ is the entropy.
Without loss of generality we may assume $r \leq s$. $H(\imath)
\leq \ln r$ with equality iff $\t_{i\p} = {1\over r}$ for all $i$.
$H(\imath|\jmath) \geq 0$ with equality iff $\imath$ is a
deterministic function of $\jmath$. Together, $I(\tilde{\v\t}) =
I_{max}$ iff $\tilde\t_{ij}={1\over
r}\delta_{i,m(j)}\cdot\sigma_j$, where $m:\{1...s\}\to\{1..r\}$ is
any onto map and the $\sigma_j\geq 0$ respect the constraints
$\sum_{j \in m^{-1}(i)}\sigma_j=1$. This suggests the
reparametrization $\t_{ij} = {1\over r}\delta_{i,m(j)}\sigma_j +
\Delta_{ij}$ in the integral (\ref{midistr}) for each choice of
$m()$ and suitable constraints on $\sigma$ and $\Delta$.

\section{Numerics}\label{secNum}
In order to approximate the distribution of mutual information in
practice, one needs consider implementation issues and the
computational complexity of the overall method. This is what we
set out to do in the following.

\subsection{Computational complexity and accuracy}
Regarding computational complexity, there are short and fast
implementations of $\psi$. The code of the Gamma function in
\cite{Press:92}, for instance, can be modified to compute the
$\psi$ function. For integer and half-integer values one may
create a lookup table from (\ref{psi2}).
The needed quantities $J$, $K$, $L$, $M$, and $Q$ (depending on
$\v\n$) involve a double sum, $P$ only a single sum, and the $r +
s$ quantities $J_{i\p}$ and $J_{\p j}$ also only a single sum.
Hence, the computation time for the (central) moments is of the
same order $O(r \cdot s)$ as for $I(\hat{\v\t})$.

With respect to the quality of the approximation, let us briefly
consider the case of the variance. The expression for the exact
variance has been Taylor-expanded in $(\frac{rs}{n})$, so the
relative error ${\frac{\mbox{\scriptsize
Var}[I]_{approx}-\mbox{\scriptsize
Var}[I]_{exact}}{\mbox{\scriptsize Var}[I]_{exact}}}$ of the
approximation is of the order $(\frac{rs}{n})^2$, {\em if}
$\imath$ and $\jmath$ are dependent. In the opposite case, the
$O(n^{-1})$ term in the sum drops itself down to order $n^{-2}$
resulting in a reduced relative accuracy $O(\frac{rs}{n})$ of the
approximated variance. These results were confirmed by numerical
experiments that we realized by Monte Carlo simulation to obtain
``exact'' values of the variance for representative choices of
$\t_{ij}$, $r$, $s$, and $n$. The approximation for the variance,
together with those for the skewness and kurtosis, and the exact
expression for the mean, allow a good description of the
distribution $p(I|\v\n)$ to be obtained for not too small sample
bin sizes $n_{ij}$.

We want to conclude with some notes on {\it useful} accuracy. The
hypothetical prior sample sizes $\n''_{ij}=\{0,{1\over
rs},\odt,1\}$ can all be argued to be non-informative
\cite{Gelman:95}. Since the central moments are expansions in
$\npp^{-1}$, the second leading order term can be freely adjusted
by adjusting $\n''_{ij}\in[0...1]$.
So one may argue that anything beyond the leading order term is
free to will, and the leading order terms may be regarded as
accurate as we can specify our prior knowledge. On the other hand,
exact expressions have the advantage of being safe against
cancellations. For instance, the leading orders of $E[I]$ and
$E[I^2]$ do not suffice to compute the leading order term of
$\Var[I]$.

\subsection{Approximating the distribution}\label{BA}
Let us now consider approximating the overall distribution of
mutual information based on the mean and the variance. Fitting a
normal distribution is an obvious possible choice, as the central
limit theorem ensures that $p(I|\v n)$ converges to a Gaussian
distribution with mean $E[I]$ and variance $\Var[I]$. Since $I$ is
non-negative, it is also worth considering the approximation of
$p(I|\v\t)$ by a Gamma (i.e., a scaled $\chi^{2}$). Another
natural candidate is the Beta distribution, which is defined for
variables in the $[0,1]$ real interval. $I$ can be made such a
variable by a simple normalization. Of course the Gamma and the
Beta are asymptotically correct, too.

\begin{figure}[tbh]
\centerline{\includegraphics[width=1.0\textwidth]{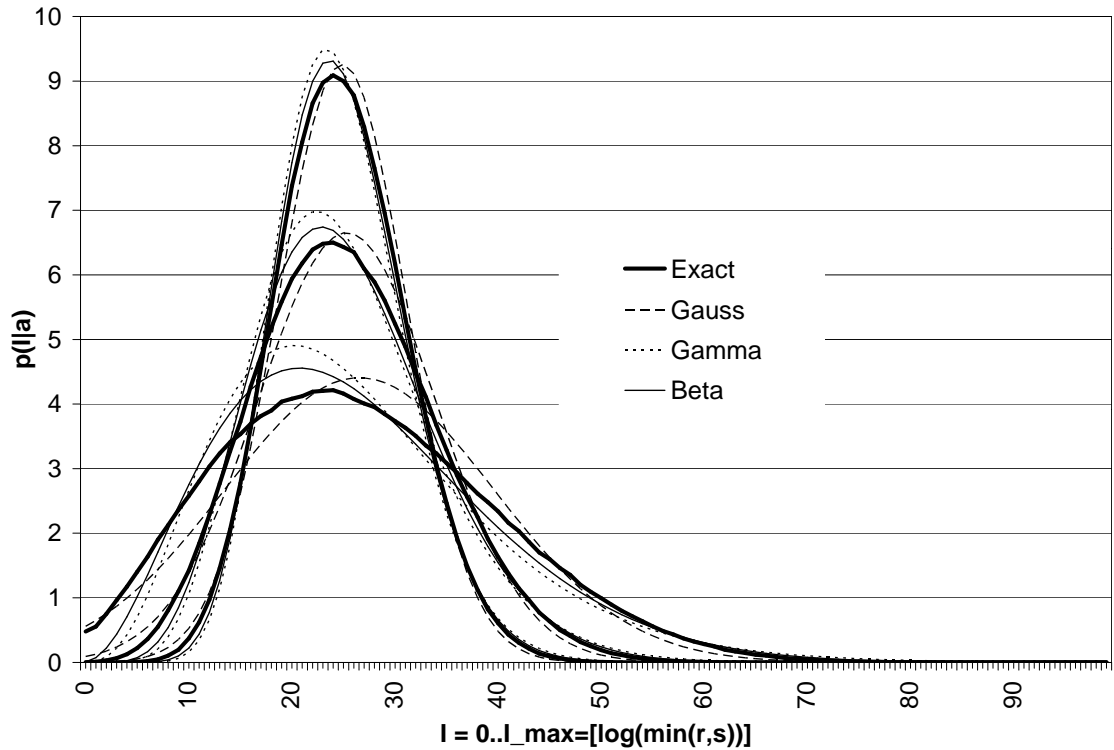}}
\caption{\label{fig1}\it Distribution of mutual
information for two binary random variables (The labelling of the
horizontal axis is the percentage of I\_$\max $.) There are three
groups of curves, for different choices of counts
$(n_{11},n_{12},n_{21},n_{22})$. The upper group is related to the
vector $(40,10,20,80)$, the intermediate one to the vector
$(20,5,10,40)$, and the lower group to $(8,2,4,16)$. Each group
shows the ``exact'' distribution and three approximating curves,
based on the Gaussian, Gamma and Beta distributions.}
\end{figure}

We report a graphical comparison of the different approximations
by focusing on the special case of binary random variables, and on
three possible vectors of counts. Figure \ref{fig1} compares the
``exact'' distribution of mutual information, computed via Monte
Carlo simulation, with the approximating curves. These curves have
been fitted using the exact mean and the approximated variance of
the preceding section. The figure clearly shows that all the
approximations are rather good, with a slight preference for the
Beta approximation. The curves tend to do worse for smaller sample
sizes, as expected. Higher moments may be used to improve the
accuracy (Section \ref{secGeneral}), or this can be improved using
our considerations about tail approximations in Section
\ref{secLASIA}.

\section{Expressions for Missing Data}\label{secMD}

In the following we generalize the setup to include the case of
missing data, which often occurs in practice.
We extend the counts $n_{ij}$ to include $n_{?j}$, which counts
the number of instances in which only $\jmath$ is observed (i.e.,
the number of $(?,j)$ instances), and the counts $n_{i?}$ for the
number of $(i,?)$ instances, where only $\imath$ is observed.

We make the common assumption that the missing data mechanism is
ignorable (\emph{missing at random} and \emph{distinct})
\cite{LitRub87}. The probability distribution of $\jmath$ given
that $\imath$ is missing coincides with the marginal $\t_{\p j}$,
and vice versa, as a consequence of this assumption.

\subsection{Setup}
The sample size $n$ is now $n_c+n_{\p?}+n_{?\p}$, where $n_c$ is
the number of complete units. Let $\v n=(n_{ij},n_{i?},n_{?j})$
denote as before the vector of counts, now including the counts
$n_{i?}$ and $n_{?j}$, for all $i$ and $j$. The probability of a
specific data set $\v D$, given $\v\t$, hence, is $p(\v
D|\v\t,n_c,n_{\p?},n_{?\p})=\prod_{ij}\t_{ij}^{n_{ij}}
\prod_i\t_{i\p}^{n_{i?}} \prod_j\t_{\p j}^{n_{?j}}$. Assuming a
uniform prior $p(\v\t)\propto 1\cdot\delta(\t_\pp-1)$, Bayes' rule
leads to the posterior (which is also the likelihood in case of
uniform prior)
\beqn
  p(\v\t|\v n) = {1\over{\cal N}(\v n)}
  \prod_{ij}\t_{ij}^{n_{ij}} \prod_i\t_{i\p}^{n_{i?}}
  \prod_j\t_{\p j}^{n_{?j}}\delta(\t_\pp-1)
\eeqn
where the normalization $\cal N$ is chosen such that $\int
p(\v\t|\v n)d^{rs}\v\t=1$. With missing data there is, in general,
no closed form expression for $\cal N$ any more (cf.\ (\ref{norm})).

In the following, we restrict ourselves to a discussion of leading
order (in $n^{-1}$) expressions. In leading order, any Dirichlet
prior with $n''_{ij}=O(1)$ leads to the same results, hence we can
simply assume a uniform prior. In leading order, the mean $E[\v\t]$
coincides with the mode of $p(\v\t|\v n)$, i.e. the maximum
likelihood estimate of $\v\t$. The log-likelihood function
$\ln\,p(\v\t|\v n)$ is
\beqn
  L(\v\t|\v n) = \sum_{ij}n_{ij}\ln\t_{ij}
  + \sum_i n_{i?}\ln\t_{i\p} + \sum_j n_{?j}\ln\t_{\p j}
  - \ln {\cal N}(\v n) - \lambda(\t_\pp-1),
\eeqn
where we have introduced the Lagrange multiplier $\lambda$ to take
into account the restriction $\t_\pp=1$. The maximum is at
${\partial L\over\partial\t_{ij}} = {n_{ij}\over\t_{ij}} +
{n_{i?}\over\t_{i\p}} + {n_{?j}\over\t_{\p j}} - \lambda = 0$.
Multiplying this by $\t_{ij}$ and summing over $i$ and $j$ we
obtain $\lambda=n$.
The maximum likelihood estimate $\hat{\v\t}$ is, hence, given by
\beq\label{EM}
  \hat\t_{ij}={1\over n}\left(n_{ij}+n_{i?}{\hat\t_{ij}\over\hat\t_{i\p}}
  + n_{?j}{\hat\t_{ij}\over\hat\t_{\p j}}\right).
\eeq
This is a non-linear equation in $\hat{\t}_{ij}$, which, in
general, has no closed form solution. Nevertheless Eq. (\ref{EM})
can be used to approximate $\hat{\t}_{ij}$. Eq.\ (\ref{EM})
coincides with the popular expectation-maximization (EM) algorithm
\cite{Chen:74} if one inserts a first estimate
$\hat{\t}_{ij}^{0}={\frac{n_{ij}}{n}}$ into the r.h.s.\ of (\ref
{EM}) and then uses the resulting l.h.s.\ $\hat{\t}_{ij}^{1}$ as a
new estimate, etc. This iteration (quickly) converges to the maximum
likelihood solution (if missing instances are not too frequent).
Using this we can compute the leading order term for the mean of
the mutual information (and of any other function of $\t_{ij}$):
$E[I]=I(\hat{\v\t})+O(n^{-1})$. The leading order term for the
covariance can be obtained from the second derivative of $L$.

\subsection{Unimodality of $p(\v\t|\v n)$}
The $rs\times rs$ Hessian matrix $\v H\in\SetR^{rs\cdot rs}$ of
$-L$ and the second derivative in the direction of the
$rs$-dimensional column vector $\v v\in\SetR^{rs}$ are
\beqn
  \v H_{(ij)(kl)}[\v\t] \;:=\;
  -{\partial L\over\partial\t_{ij}\partial\t_{kl}}
  = {n_{ij}\over\t_{ij}^2}\delta_{ik}\delta_{jl} +
    {n_{i?}\over\t_{i\p}^2}\delta_{ik} +
    {n_{?j}\over\t_{\p j}^2}\delta_{jl},
\eeqn
\beqn
  \v v^T\v H\v v \;=\;
  \sum_{ijkl}v_{ij} \v H_{(ij)(kl)}v_{kl} \;=\;
  \sum_{ij}{n_{ij}\over\t_{ij}^2}v_{ij}^2 +
  \sum_i{n_{i?}\over\t_{i\p}^2}v_{i\p}^2 +
  \sum_j{n_{?j}\over\t_{\p j}^2}v_{\p j}^2 \;\geq\; 0.
\eeqn
This shows that $-L$ is a convex function of $\v\t$, hence
$p(\v\t|\v n)$ has a single (possibly degenerate) global maximum.
$L$ is strictly convex if $n_{ij}>0$ for all $ij$, since $\v v^T\v
H\v v>0$ $\forall\,\v v\neq 0$ in this case. (Note that positivity
of $n_{i?}$ for all $i$ is not sufficient, since $v_{i\p}=0$ for
$\v v\neq 0$ is possible. Actually $v_\pp=0$.) This implies a
unique global maximum, which is attained in the interior of the
probability simplex. Since EM is known to converge to a local
maximum, this shows that in fact {\em EM always converges to the
global maximum}.

\subsection{Covariance of $\v\t$}
With
\beqn
  \v A_{(ij)(kl)} := \v H_{(ij)(kl)}[\hat{\v\t}]
  = n\left[{\delta_{ik}\delta_{jl}\over\rho_{ij}} \!+\!
    {\delta_{ik}\over\rho_{i?}} \!+\!
    {\delta_{jl}\over\rho_{?j}}\right],
\eeqn
\beq\label{kernelA}
    \rho_{ij}:=n{\hat\t_{ij}^2\over n_{ij}},\quad
    \rho_{i?}:=n{\hat\t_{i\p}^2\over n_{i?}},\quad
    \rho_{?j}:=n{\hat\t_{\p j}^2\over n_{?j}}
\eeq
and
$\v\Delta:=\v\t-\hat{\v\t}$, we can
represent the posterior to leading order as an $(rs-1)$-dimensional
Gaussian:
\beq\label{postgauss}
  p(\v\t|\v n) \sim e^{-{1\over 2}\v\Delta^T\v A\v\Delta}
  \delta(\Delta_\pp).
\eeq
The easiest way to compute the covariance (and other quantities)
is to also represent the $\delta$-function as a narrow Gaussian of
width $\eps\approx 0$. Inserting $\delta(\Delta_\pp)\approx
{1\over\eps\sqrt{2\t}} \exp(-{1\over 2\eps^2}\v\Delta^T\v e\v
e^T\v\Delta)$ into (\ref{postgauss}), where $\v e_{ij}=1$ for all
$ij$ (hence $\v e^T\v\Delta=\Delta_\pp$), leads to a full
$rs$-dimensional Gaussian with kernel $\tilde{\v A}=\v A+\v u\v
v^T$, $\v u=\v v={1\over\eps}\v e$. The covariance of a Gaussian
with kernel $\tilde{\v A}$ is $\tilde{\v A}^{-1}$. Using the
Sherman-Morrison formula $\tilde{\v A}^{-1}=\v A^{-1}-\v A^{-1}{\v
u\v v^T\over 1+\v v^T\v A^{-1}\v u}\v A^{-1}$
\cite[p.~73]{Press:92} and $\eps\to 0$ we get
\beq\label{covmis}
  \Cov_{(ij)(kl)}[\v\t] :=
  E[\Delta_{ij}\Delta_{kl}] \eqsq
  [\tilde{\v A}^{-1}]_{(ij)(kl)} =
  \left[\v A^{-1} - {\v A^{-1}\v e\v e^T\v
  A^{-1}\over\v e^T\v A^{-1}\v e}\right]_{(ij)(kl)},
\eeq
where $\eqsq$ denotes equality up to terms of order $n^{-2}$.
Singular matrices $\v A$ are easily avoided by choosing a prior
such that $n_{ij}>0$ for all $i$ and $j$. $\v A$ may be inverted
exactly or iteratively, the latter by a trivial inversion of the
diagonal part $\delta_{ik}\delta_{jl}/\rho_{ij}$ and by treating
$\delta_{ik}/\rho_{i?} + \delta_{jl}/\rho_{?j}$ as a perturbation.

\subsection{Missing observations for one variable only}
In the case only one variable is missing, say
$n_{?j}=0$, closed form expressions can be obtained. If we sum
(\ref{EM}) over $j$ we get $\hat\t_{i\p}={n_{i\p}+n_{i?}\over n}$.
Inserting $\hat\t_{i\p}={n_{i\p}+n_{i?}\over
n}$ into the r.h.s.\ of (\ref{EM}) and solving w.r.t.\
$\hat\t_{ij}$, we get the explicit expression
\beq\label{pimfo}
  \hat\t_{ij}={n_{i\p}\!+\!n_{i?}\over n}\cdot{n_{ij}\over n_{i\p}}.
\eeq
Furthermore, it can easily be verified (by multiplication) that
$\v A_{(ij)(kl)}=n[\delta_{ik}\delta_{jl}/\rho_{ij}+
\delta_{ik}/\rho_{i?}]$ has inverse
$
  [\v A^{-1}]_{(ij)(kl)} =
  {1\over n}[\rho_{ij}\delta_{ik}\delta_{jl} -
  {\rho_{ij}\rho_{kl}\over\rho_{i\p}\!+\!\rho_{i?}}\delta_{ik}]
$.
With the abbreviations
\beqn
  \tilde Q_{i?}:= {\rho_{i?}\over\rho_{i?}+\rho_{i\p}}
  \qmbox{and}
  \tilde Q:=\sum_i\rho_{i\p}\tilde Q_{i?}
\eeqn
we get $[\v A^{-1}\v e]_{ij} = \sum_{kl}[\v A^{-1}]_{(ij)(kl)} =
{1\over n}\rho_{ij}\tilde Q_{i?}$ and $\v e^T\v A^{-1}\v e =
\tilde Q/n$.
Inserting everything into (\ref{covmis}) we get
\beqn
  \Cov_{(ij)(kl)}[\v\t] \;\eqsq\;
  {1\over n}\left[\rho_{ij}\delta_{ik}\delta_{jl} -
  {\rho_{ij}\rho_{kl}\over\rho_{i\p}\!+\!\rho_{i?}}\delta_{ik}
  - {\rho_{ij}\tilde Q_{i?}\rho_{kl}\tilde Q_{k?}\over\tilde
  Q}\right].
\eeqn
Inserting this expression for the covariance into (\ref{varlo}),
using $\v{\mean\t}:=E[\v\t]=\hat{\v\t}+O(n^{-1})$, we finally get
the leading order term in $1/n$ for the variance of mutual
information:
\beq\label{varmfo}
  \Var[I] \;\eqsq\; {1\over n}[\tilde K-\tilde J^2/\tilde Q-\tilde P],
  \qquad\qquad
  \tilde K := \sum_{ij}\rho_{ij}
  \left(\ln{\hat\t_{ij}\over\hat\t_{i\p}\hat\t_{\p j}}\right)^2,
\eeq\vspace{-2ex}
\beqn
  \tilde P \;:=\; \sum_i{\tilde J_{i\p}^2Q_{i?}\over\rho_{i?}},
  \qquad
  \tilde J \;:=\;\sum_i \tilde J_{i\p}\tilde Q_{i?},
  \qquad
  \tilde J_{i\p}:=\sum_j\rho_{ij}\ln{\hat\t_{ij}\over\hat\t_{i\p}\hat\t_{\p j}}.
\eeqn
A closed form expression for ${\cal N}(\v n)$ also exists.
Symmetric expressions for the case when only $\imath$ is missing
can be obtained.
Note that for the complete data case $n_{i?}=0$, we have
$\hat\t_{ij}=\rho_{ij}={n_{ij}\over n}$, $\rho_{i?}=\infty$,
$\tilde Q_{i?}=\tilde Q=1$, $\tilde J=J$, $\tilde K=K$, and $\tilde P=0$,
consistent with (\ref{varlodi}).

There is at least one reason for minutely having inserted all
expressions into each other and introducing quite a number
definitions. In the presented form all expressions involve at most
a double sum. Hence, the overall time for computing the mean and
variance when only one variable is missing is $O(rs)$.

\subsection{Expressions for the general case}
In the general case when both variables are missing, each EM
iteration (\ref{EM}) for $\hat\t_{ij}$ needs $O(rs)$ operations.
The naive inversion of $\v A$ needs time $O((rs)^3)$, and using it
to compute Var[$I$] time $O((rs)^2)$. Since the contribution from
unlabelled-$\imath$ instances can be interpreted as a rank $s$
modification of $\v A$ in the case of when $\imath$ is not
missing, one can use Woodbury's formula
$
  [\v B+\v U\v D\v V^T]^{-1} =
  \v B^{-1}-\v B^{-1}\v U [\v D^{-1}+
  \v V^T\v B^{-1}\v U]^{-1}\v V^T\v B^{-1}
$ \cite[p.~75]{Press:92}
with $\v B_{(ij)(kl)}=\delta_{ik}\delta_{jl}/\rho_{ij}+
\delta_{ik}/\rho_{i?}$, $\v D_{jl}=\delta_{jl}/\rho_{?j}$, and $\v U_{(ij)l}=\v
V_{(ij)l}=\delta_{jl}$,
to reduce the inversion of the $rs\times rs$ matrix $\v A$ to the
inversion of a single $s$-dimensional matrix. The result can be
written in the form
\beq\label{AIGen}
  [\v A^{-1}]_{(ij)(kl)} = {1\over n}
  \left[F_{ijl}\delta_{ik}-\sum_{mn}F_{ijm}[\v G^{-1}]_{mn}F_{kln}\right],
\eeq
\beqn
  F_{ijl} := \rho_{ij}\delta_{jl} -
    {\rho_{ij}\rho_{kl}\over\rho_{i?}\!+\!\rho_{i+}},\qquad
  G_{mn} := \rho_{?n}\delta_{mn}+F_{\p mn}.
\eeqn
The result for the covariance (\ref{covmis}) can be inserted into
(\ref{varlo}) to obtain the leading order term for the variance:
\beq\label{VarGen}
  \Var[I] \;\eqsq\; \v l^T\v A^{-1}\v l - (\v l^T\v A^{-1}\v
  e)^2/(\v e^T\v A^{-1}\v e) \qmbox{where}
  \v l_{ij}:=\ln{\hat\t_{ij}\over\hat\t_{i\p}\hat\t_{\p j}}.
\eeq
Inserting (\ref{AIGen}) into (\ref{VarGen}) and rearranging terms
appropriately, we can compute Var[$I$] in time $O(rs)$ plus the
time $O(s^2r)$ to compute the $s\times s$ matrix $\v G$ and time
$O(s^3)$ to invert it, plus the time $O(\#\!\cdot\!rs)$ for
determining $\hat\t_{ij}$, where $\#$ is the number of iterations
of EM. Of course, one can and should always choose $s\leq r$. Note
that these expressions converge to the exact values when $n$ goes
to infinity, irrespectively of the amount of missing data.

\section{Applications}\label{sec:discussion}

The results in the preceding sections provide fast and reliable
methods to approximate the distribution of mutual information from
either complete or incomplete data. The derived tools have been
obtained in the theoretically sound framework of Bayesian
statistics, which we regard as their basic justification. As these
methods are available for the first time, it is natural to wonder
what their possible uses can be on the application side or, stated
differently, what can be gained in practice moving from
descriptive to inductive methods. We believe that the impact on
real applications can be significant, according to three main
scenarios: \emph{robust inference methods}, \emph{inferring models
that perform well}, and \emph{fast learning from massive data
sets}. In the following we use classification as a thread to
illustrate the above scenarios. \emph{Classification} is one of
the most important techniques for knowledge discovery in databases
\cite{DuHaSt01}. A classifier is an algorithm that allocates new
objects to one out of a finite set of previously defined groups
(or \emph{classes}) on the basis of observations on several
characteristics of the objects, called \emph{attributes} or
\emph{features}. Classifiers are typically learned from data,
making explicit the knowledge that is hidden in databases, and
using this knowledge to make predictions about new data.

\subsection{Robust inference methods}\label{sec:robust}
An obvious observation is that descriptive methods cannot
compete, by definition, with inductive ones when robustness is
concerned. Hence, the results presented in this paper lead
naturally to a spin-off for reliable methods of inference.

Let us focus on classification problems, for the sake of
explanation. Applying robust methods to classification means to
produce classifications that are correct with a given probability.
It is easy to imagine sensible (e.g., nuclear,
medical) applications where reliability of classification is a
critical issue. To achieve reliability, a necessary step consists
in associating a posterior probability (i.e., a guarantee level)
to classification models inferred from data, such as
classification trees or Bayesian nets. Let us consider the case of
Bayesian networks. These are graphical models that represent
structures of (in)dependence by directed acyclic graphs, where
nodes in the graph are regarded as random variables
\cite{Pearl88,Neapolitan:04}. Two nodes are connected by an arc
when there is direct stochastic dependence between them. Inferring
Bayesian nets from data is often done by connecting nodes with
significant value of descriptive mutual information. Little work
has been done on robustly inferring Bayesian nets, probably
because of the difficulty to deal with the distribution of mutual
information, with the notable exception of Kleiter's work
\cite{Kleiter:99}. Joining Kleiter's work with ours might lead to
inference of Bayesian network structures that are correct with a
given probability. Some work has already been done to this
direction \cite{zaffalon03a}.

Feature selection might also benefit from robust methods.
\emph{Feature selection} is the problem of reducing the number of
feature variables to deal with in classification problems.
Features can reliably be discarded only when they are irrelevant
to the class with high probability. This needs knowledge of the
distribution of mutual information. In Section \ref{secFS} we propose
a filter based on the distribution of mutual information to
address this problem.

\subsection{Inferring models that perform well}\label{sec:well}
It is well-known that model complexity must be in proper balance
with available data in order to achieve good classification
accuracy. In fact, unjustified complexity of inferred models leads
classifiers almost inevitably to \emph{overfitting}, i.e.\ to
memorize the available sample rather than extracting regularities
from it that are needed to make useful predictions on new data
\cite{DuHaSt01}. Overfitting could be avoided by using the
distribution of mutual information. With Bayesian nets, for
example, this could be achieved by drawing arcs between nodes only
if these are supported by data with high probability. This is a
way to impose a bias towards simple structures. It has to be
verified whether or not this approach can systematically lead to
better accuracy.

Model complexity can also be reduced by discarding features. This
can be achieved by including a feature only when its mutual
information with the class is significant with high probability.
This approach is taken in Section \ref{secFS}, where we show that it
can effectively lead to better prediction accuracy of the
resulting models.

\subsection{Fast learning from massive data sets}\label{sec:massive}
Another very promising application of the distribution of mutual
information is related to \emph{massive data sets}. These are huge
samples, which are becoming more and more available in real
applications, and which constitute a serious challenge for machine
learning and statistical applications. With massive data sets it
is impractical to scan all the data, so classifiers must be
reliably inferred by accessing only a small subset of the units.
Recent work has highlighted \cite{pelleg2003} how inductive
methods allow this to be realized. The intuition is the following:
the inference phase stops reading data when the inferred model,
say a Bayesian net, has reached a given posterior probability. By
choosing such probability sufficiently high, one can be
arbitrarily confident that the inferred model will not
change much by reading the neglected data, making the
remaining units superfluous.

\section{Feature Selection}\label{secFS}

Feature selection is a basic step in the process of building
classifiers \cite{BluLan97,DasLiu97,LiuMot98}. In fact, even if
theoretically more features should provide one with better
\emph{prediction accuracy} (i.e., the relative number of correct
predictions), in real cases it has been observed many times that
this is not the case \cite{KolSah96} and that it is important to
discard irrelevant, or weakly relevant features.

The purpose of this section is to illustrate how the distribution
of mutual information can be applied in this framework, according
to some of the ideas in Section \ref{sec:well}. Our goal is
inferring simple models that avoid overfitting and have an
equivalent or better accuracy with respect to models that consider
all the original features.

Two major approaches to feature selection are commonly used in
machine learning \cite{JohKohPfl94}: \emph{filter} and
\emph{wrapper} models. The filter approach is a preprocessing step
of the classification task. The wrapper model is computationally
heavier, as it implements a search in the feature space using the
prediction accuracy as reward measure. In the following we focus
our attention on the filter approach: we define two new filters
and report experimental analysis about them, both with complete
and incomplete data.

\subsection{The proposed filters}\label{TPF}
We consider the well-known filter (F) that computes the empirical
mutual information between features and the class, and discards
low-valued features \cite{Lew92}. This is an easy and effective
approach that has gained popularity with time. Cheng reports that
it is particularly well suited to jointly work with Bayesian
network classifiers, an approach by which he won the \emph{2001
international knowledge discovery competition}
\cite{CheHatHayKroMorPagSes02}. The `Weka' data mining package
implements it as a standard system tool (see \cite[p.~294]{WitFra99}).

A problem with this filter is the variability of the empirical
mutual information with the sample. This may cause wrong judgments
of relevance, when those features are selected
for which the mutual information exceeds a fixed threshold
$\eps$. In order for the selection to be robust, we must
have some guarantee about the actual value of mutual information.

We define two new filters. The \emph{backward filter }(BF) {\em
discards} an attribute if $p(I<\eps|\v n)>\bar p$ where $I$
denotes the mutual information between the feature and the class,
$\eps$ is an arbitrary (low) positive threshold and $\bar p$ is an
arbitrary (high) probability. The \emph{forward filter} (FF) {\em
includes} an attribute if $p(I>\eps|\v n)>\bar p$, with the same
notations. BF is a conservative filter, along the lines discussed
about robustness in Section \ref{sec:robust}, because it will only
discard features after observing substantial evidence supporting
their irrelevance. FF instead will tend to use fewer features
(aiming at producing classifiers that perform better), i.e. only
those for which there is substantial evidence about them being
useful in predicting the class.

The next sections present experimental comparisons of the new
filters and the original filter F.

\subsection{Experimental methodology}\label{secEA}
For the following experiments we use the \emph{naive Bayes classifier}
\cite{DuHa73}. This is a good classification model---despite its
simplifying assumptions \cite{DoPa97}---, which often
competes successfully with much more complex classifiers from the
machine learning field, such as C4.5 \cite{Quinlan93}. The
experiments focus on the incremental use of the naive Bayes
classifier, a natural learning process when the data are available
sequentially: the data set is read instance by instance; each
time, the chosen filter selects a subset of attributes that the
naive Bayes uses to classify the new instance; the naive Bayes
then updates its knowledge by taking into consideration the new
instance and its actual class. The incremental approach allows us
to better highlight the different behaviors of the empirical
filter (F) and those based on the distribution of mutual
information (BF and FF). In fact, for increasing sizes of the
learning set the filters converge to the same behavior.

For each filter, we are interested in experimentally evaluating
two quantities: for each instance of the data set, the average
number of correct predictions (namely, the prediction accuracy) of
the naive Bayes classifier up to such instance; and the average
number of attributes used. By these quantities we can compare the
filters and judge their effectiveness.

The implementation details for the following experiments include:
using the Beta approximation (Section \ref{BA}) to the
distribution of mutual information, with the exact mean
(\ref{miexex}) and the $O\left( n^{-3}\right) $-approximation of
the variance, given in (\ref{var2ndo}); using the uniform prior
for the naive Bayes classifier and all the filters; and setting
the level $\bar p$ for the posterior probability to $0.95$. As far
as $\eps $ is concerned, we cannot set it to zero because the
probability that two variables are independent ($I=0$) is zero
according to the inferential Bayesian approach. We can interpret
the parameter $\eps $ as a degree of dependency strength below
which attributes are deemed irrelevant. We set $\eps $ to $0.003$,
in the attempt of only discarding attributes with negligible
impact on predictions. As we will see, such a low threshold can
nevertheless bring to discard many attributes.

\section{Experimental analysis with incomplete samples}\label{DS}

Table \ref{tab1} lists ten data sets used in the experiments for
complete data. These are real data sets on a number of different
domains. For example, Shuttle-small reports data on diagnosing
failures of the space shuttle; Lymphography and Hypothyroid are
medical data sets; Spam is a body of e-mails that can be spam or
non-spam; etc.

\begin{table}[h]
\begin{minipage}{0.6\textwidth}\tabcolsep=1ex
\begin{tabular}{lrrr}
\hline
Name & \#feat. & \#inst.& mode freq. \\ \hline
Australian &  36 &  690 & 0.555 \\
Chess      &  36 & 3196 & 0.520 \\
Crx        &  15 &  653 & 0.547 \\
German-org &  17 & 1000 & 0.700 \\
Hypothyroid & 23 & 2238 & 0.942 \\
Led24      &  24 & 3200 & 0.105 \\
Lymphography & 18 & 148 & 0.547 \\
Shuttle-small & 8 & 5800 & 0.787 \\
Spam     & 21611 & 1101 & 0.563 \\
Vote       &  16 &  435 & 0.614 \\ \hline
\end{tabular}
\end{minipage}
\begin{minipage}{0.4\textwidth}\vspace{-4ex}
\caption{\label{tab1}\it Complete data sets used
in the experiments, together with their number of features, of
instances and the relative frequency of the mode. All but the Spam
data sets are available from the UCI repository of machine
learning data sets \cite{MurAha95}. The Spam data set is described
in \cite{AndKouKonChaPalSpy00} and available from
Androutsopoulos's web page.}
\end{minipage}
\end{table}

The data sets presenting non-categorical features have been
pre-discretized by MLC++ \cite{KoJoLoMaPf94}, default options,
i.e.\ by the common entropy-based discretization \cite{FayIra93}.
This step may remove some attributes judging them as irrelevant,
so the number of features in the table refers to the data sets
after the possible discretization. The instances with missing
values have been discarded, and the third column in the table
refers to the data sets without missing values. Finally, the
instances have been randomly sorted before starting the
experiments.

\subsection{Results}\label{R}
In short, the results show that FF outperforms the commonly used
filter F, which in turn, outperforms the filter BF. FF leads
either to the same prediction accuracy as F or to a better one,
using substantially fewer attributes most of the times. The same
holds for F versus BF.

In particular, we used the \emph{two-tails paired t test} at level
0.05 to compare the prediction accuracies of the naive Bayes with
different filters, in the first $k$ instances of the data set, for
each $k$.

The results in Table \ref{tab2} show that, despite the number of
used attributes is often substantially different, both the
differences between FF and F, and the differences between F and
BF, were never statistically significant on eight data sets out of
ten.

\begin{table}[h]
\begin{minipage}{0.6\textwidth}\tabcolsep=1ex
\begin{tabular}{lrrrr}
\hline\small
Data set & \hspace{-3ex}\#feat. &   FF &    F &   BF \\ \hline
Australian     & 36 & 32.6 & 34.3 & 35.9 \\
\textbf{Chess} & 36 & 12.6 & 18.1 & 26.1 \\
Crx            & 15 & 11.9 & 13.2 & 15.0 \\
German-org     & 17 &  5.1 &  8.8 & 15.2 \\
Hypothyroid    & 23 &  4.8 &  8.4 & 17.1 \\
Led24          & 24 & 13.6 & 14.0 & 24.0 \\
Lymphography   & 18 & 18.0 & 18.0 & 18.0 \\
Shuttle-small  &  8 &  7.1 &  7.7 &  8.0 \\
\textbf{Spam}&\hspace{-3ex}21611 &123.1 &822.0 &\hspace{-1ex}13127.4 \\
Vote           & 16 & 14.0 & 15.2 & 16.0 \\ \hline
\end{tabular}
\end{minipage}
\begin{minipage}{0.4\textwidth}\vspace{-0ex}
\caption{\label{tab2}\it Average number of
attributes selected by the filters on the entire data set,
reported in the last three columns. (Refer to the Section `The
proposed filters' for the definition of the filters.) The second
column from left reports the original number of features. In all
but one case, FF selected fewer features than F, sometimes much
fewer; F usually selected much fewer features than BF, which was
very conservative. Boldface names refer to data sets on which
prediction accuracies where significantly different.}
\end{minipage}\hspace{5ex}
\end{table}

\begin{figure}[tbh]
\begin{minipage}{0.6\textwidth}
\includegraphics[width=0.9\textwidth]{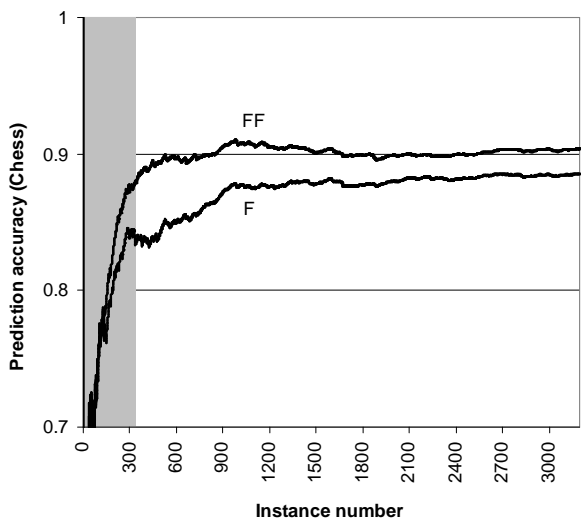}
\end{minipage}
\begin{minipage}{0.4\textwidth}\vspace{-4ex}
\caption{\label{fig2}\it Comparison of the
prediction accuracies of the naive Bayes with filters F and FF on
the Chess data set. The gray area denotes differences that are not
statistically significant.}
\end{minipage}
\end{figure}

The remaining cases are described by means of the following
figures. Figure \ref{fig2} shows that FF allowed the naive Bayes
to significantly do better predictions than F for the greatest
part of the Chess data set. The maximum difference in prediction
accuracy is obtained at instance 422, where the accuracies are
0.889 and 0.832 for the cases FF and F, respectively. Figure
\ref{fig2} does not report the BF case, because there is no
significant difference with the F curve. The good performance of
FF was obtained using only about one third of the attributes
(Table \ref{tab2}).

\begin{figure}[tbh]
\begin{minipage}{0.6\textwidth}
\includegraphics[width=0.9\textwidth]{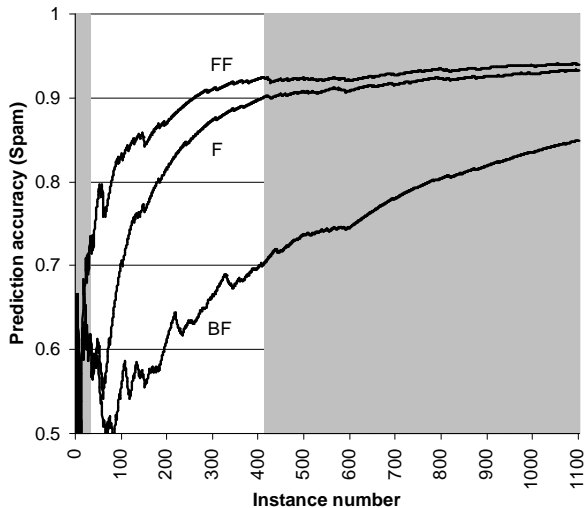}
\end{minipage}
\begin{minipage}{0.4\textwidth}\vspace{-4ex}
\caption{\label{fig3}\it Prediction accuracies
of the naive Bayes with filters F, FF and BF on the Spam data set.
The differences between F and FF are significant in the range of
observations 32--413. The differences between F and BF are
significant from observations 65 to the end (this significance is
not displayed in the picture).}
\end{minipage}
\end{figure}

\begin{figure}[tbh]
\begin{minipage}{0.6\textwidth}
\includegraphics[width=0.9\textwidth]{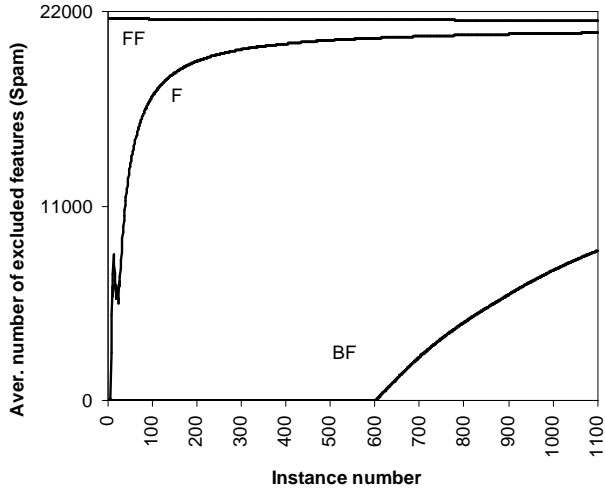}
\end{minipage}
\begin{minipage}{0.4\textwidth}\vspace{-4ex}
\caption{\label{fig4}\it Average number of attributes
excluded by the different filters on the Spam data set.}
\end{minipage}
\end{figure}

Figure \ref{fig3} compares the accuracies on the Spam data set.
The difference between the cases FF and F is significant in the
range of instances 32--413, with a maximum at instance 59 where
accuracies are 0.797 and 0.559 for FF and F, respectively. BF is
significantly worse than F from instance 65 to the end. This
excellent performance of FF is even more valuable considered the
very low number of attributes selected for classification. In the
Spam case, attributes are binary and correspond to the presence or
absence of words in an e-mail and the goal is to decide whether or
not the e-mail is spam. All the 21611 words found in the body of
e-mails were initially considered. FF shows that only an average
of about 123 relevant words is needed to make good predictions.
Worse predictions are made using F and BF, which select, on
average, about 822 and 13127 words, respectively. Figure
\ref{fig4} shows the average number of excluded features for the
three filters on the Spam data set. FF suddenly discards most of
the features, and keeps the number of selected features almost
constant over all the process. The remaining filters tend to such
a number, with different speeds, after initially including many
more features than FF.

In summary, the experimental evidence supports the strategy of
only using the features that are reliably judged to carry useful
information to predict the class, provided that the judgment can
be updated as soon as new observations are collected. FF almost
always selects fewer features than F, leading to a prediction
accuracy at least as good as the one F leads to. The comparison
between F and BF is analogous, so FF appears to be the best filter
and BF the worst. This is not surprising as BF was designed to be
conservative and was used here just as a term of comparison. The
natural use of BF is for robust classification when it is
important not to discard features potentially relevant to predict
the class.

\section{Experimental analysis with incomplete samples}\label{EAWIS}
This section makes experimental analysis on incomplete data along
the lines of the preceding experiments. The new data sets are
listed in Table \ref{tab3}.

\begin{table}[h]
\begin{minipage}{0.6\textwidth}\tabcolsep=0.5ex
\begin{tabular}{lrrrr}
\hline\small
Name & \hspace{-3ex}\#feat. & \#inst. & \#m.d. & mode freq.\\ \hline
Audiology       & 69 & 226 &  317 & 0.212 \\
Crx             & 15 & 690 &   67 & 0.555 \\
Horse-colic     & 18 & 368 & 1281 & 0.630 \\
Hypothyroidloss & 23 &3163 & 1980 & 0.952 \\
Soybean-large   & 35 & 683 & 2337 & 0.135 \\ \hline
\end{tabular}
\end{minipage}
\begin{minipage}{0.4\textwidth}\vspace{-0ex}
\caption{\label{tab3}\it Incomplete data sets
used for the new experiments, together with their number of
features, instances, missing values, and the relative frequency of
the mode. The data sets are available from the UCI repository of
machine learning data sets \cite{MurAha95}.}
\end{minipage}
\end{table}%

The filters F and FF are defined as before. However, now the mean
and variance of mutual information are obtained by using the
results in Section \ref{secMD}, in particular the closed-form
expressions for the case when only one variable is missing. In
fact, in the present data sets the class is never missing, as it
is quite common classification tasks. We remark that the mean is
simply approximated now as $I(\hat{\v\t})$, where $\hat{\v\t}$ is
given by (\ref{pimfo}), whereas the variance is reported in
(\ref{varmfo}). Furthermore, note that also the traditional filter
F, as well as the naive Bayes classifier, are now computed using
the empirical probabilities (\ref{pimfo}). The remaining
implementation details are as in the case of complete data.

\begin{table}[h]
\begin{minipage}{0.6\textwidth}\tabcolsep=1ex
\begin{tabular}{lrrrr}
\hline
Data set  & \hspace{-3ex}\#feat. &   FF &    F &   BF \\ \hline
Audiology &       69 & 64.3 & 68.0 & 68.7 \\
Crx       &       15 &  9.7 & 12.6 & 13.8 \\
Horse-colic &     18 & 11.8 & 16.1 & 17.4 \\
\textbf{Hypothyroidloss} & 23 & 4.3 & 8.3 & 13.2 \\
Soybean-large &   35 & 34.2 & 35.0 & 35.0 \\ \hline
\end{tabular}
\end{minipage}
\begin{minipage}{0.4\textwidth}\vspace{-0ex}
\caption{\label{tab4}\it Average number of
attributes selected by the filters on the entire data set,
reported in the last three columns. The second column from left
reports the original number of featurs. FF always selected fewer
features than F; F almost always selected fewer features than BF.
Prediction accuracies where significantly different for the
Hypothyroidloss data set.}
\end{minipage}
\end{table}

\begin{figure}[tbh]
\begin{minipage}{0.6\textwidth}
\includegraphics[width=0.9\textwidth]{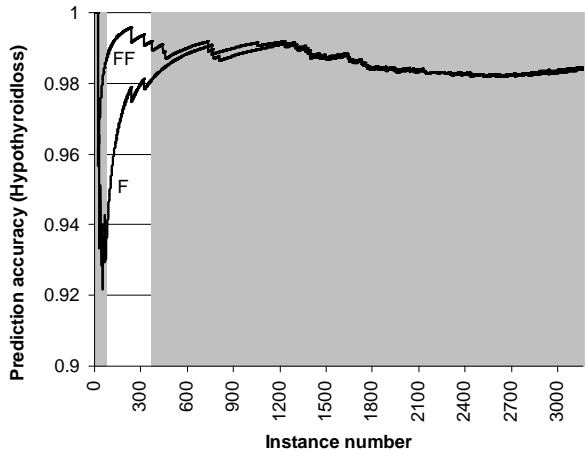}
\end{minipage}
\begin{minipage}{0.4\textwidth}\vspace{-4ex}
\caption{\label{fig5}\it Prediction accuracies
of the naive Bayes with filters F and FF on the Hypothyroidloss
data set. (BF is not reported because there is no significant
difference with the F curve.) The differences between F and FF are
significant in the range of observations 71--374. The maximum
difference is achieved at observation 71, where the accuracies are
0.986 (FF)\ vs. 0.930 (F).}
\end{minipage}
\end{figure}

The results in Table \ref{tab4} show that the filters behave very
similarly to the case of complete data. The filter FF still
selects the smallest number of features, and this number usually
increases with F and even more with BF. The selection can be very
pronounced, as with the Hypothyroidloss data set. This is also the
only data set for which the prediction accuracies of F and FF are
significantly different, in favor of FF. This is better
highlighted by Figure \ref{fig5}.

\vspace{4ex}

\subsection{Remark}
The most prominent evidence from the experiments is the better
performance of FF versus the traditional filter F. In this note
we look at FF from another perspective to exemplify and explain
its behavior.

FF includes an attribute if $p(I>\eps|\v n)>\bar p$,
according to its definition. Let us assume that FF is realized by
means of the Gaussian rather than the Beta approximation (as in
the experiments above), and let us choose $\bar p\approx 0.977$.
The condition $p(I>\eps|\v n)>\bar p$ becomes $\eps
<E[I]-2\cdot \sqrt{\Var[I]}$, or, in an approximate way,
$I(\hat{\v\t})>\eps+2\cdot \sqrt{\Var[I]}$, given that
$I(\hat{\v\t})$ is the first-order approximation of $E[I]$ (cf.
\eqref{exnlo}). We can regard $\eps+2\cdot \sqrt{\Var[I]}$
as a new threshold $\eps'$. Under this interpretation, we
see that FF is approximately equal to using the filter F with the
bigger threshold $\eps'$. This interpretation makes it also
clearer why FF can be better suited than F for sequential learning
tasks. In sequential learning, $\Var[I]$ decreases as new units
are read; this makes $\eps'$ a self-adapting
threshold that adjusts the level of caution (in including
features) as more units are read. In the limit, $\eps'$ is
equal to $\eps$. This characteristic of self-adaptation,
which is absent in F, seems to be decisive to the success of FF.

\section{Conclusions}\label{C}
This paper has provided fast and reliable analytical
approximations for the variance, skewness and kurtosis of the
posterior distribution of mutual information, with guaranteed
accuracy from $O(n^{-1})$ to $O(n^{-3})$, as well as the exact
expression of the mean. These results allow the posterior
distribution of mutual information to be approximated both from
complete and incomplete data. As an example, this paper has shown
that good approximations can be obtained by fitting common curves
with the mentioned mean and variance. To our knowledge, this is
the first work that addresses the analytical approximation of the
distribution of mutual information. Analytical approximations are
important because their implementation is shown to lead to
computations of the same order of complexity as needed for the
empirical mutual information. This makes the inductive approach a
serious competitor of the descriptive use of mutual information
for many applications.

In fact, many applications are based on descriptive mutual
information. We have discussed how many of these could benefit
from moving to the inductive side, and in particular we have shown
how this can be done for feature selection. In this context, we
have proposed the new filter FF, which is shown to be more
effective for sequential learning tasks than the traditional
filter based on empirical mutual information.


\addcontentsline{toc}{subsubsection}{References}
\begin{small}
\newcommand{\etalchar}[1]{$^{#1}$}

\end{small}

\end{document}